\documentclass[conference]{IEEEtran}
\usepackage{times}

\usepackage[numbers]{natbib}
\usepackage{multicol}
\usepackage[bookmarks=true, hidelinks]{hyperref}

\usepackage{comment}
\usepackage{siunitx}
\usepackage{relsize}
\usepackage{ifthen}
\usepackage[colorinlistoftodos]{todonotes}

\usepackage[caption=false]{subfig}

\usepackage[vlined,ruled,linesnumbered]{algorithm2e}
\usepackage{graphics} 
\usepackage{rotating}
\usepackage{color}
\usepackage{enumerate}
\usepackage[T1]{fontenc}
\usepackage{psfrag}
\usepackage{epsfig} 
\usepackage{booktabs}
\usepackage{graphicx,url}
\usepackage{multirow}
\usepackage{array}
\usepackage{latexsym}
\usepackage{amsfonts}
\usepackage{amsmath}
\usepackage{amssymb}
\usepackage{mathtools}
\usepackage{amsthm}
\usepackage{xstring}
\usepackage[noend]{algorithmic}
\usepackage{multirow}
\usepackage{xcolor}
\usepackage{prettyref}
\usepackage{flexisym}
\usepackage{bigdelim}
\usepackage{breqn} 
\usepackage{listings}
\let\labelindent\relax
\usepackage{enumitem}
\usepackage{xspace}
\usepackage{bm}
\graphicspath{{./figures/}}
\usepackage{tikz}
\usetikzlibrary{matrix,calc}
\usepackage[acronym]{glossaries}
\usepackage[capitalize]{cleveref}

\usepackage{mdwlist}

\let\itemize\undefined
\makecompactlist{itemize}{stditemize}

\newrefformat{prob}{Problem\,\ref{#1}}
\newrefformat{def}{Definition\,\ref{#1}}
\newrefformat{sec}{Section\,\ref{#1}}
\newrefformat{sub}{Section\,\ref{#1}}
\newrefformat{prop}{Proposition\,\ref{#1}}
\newrefformat{app}{Appendix\,\ref{#1}}
\newrefformat{alg}{Algorithm\,\ref{#1}}
\newrefformat{cor}{Corollary\,\ref{#1}}
\newrefformat{thm}{Theorem\,\ref{#1}}
\newrefformat{lem}{Lemma\,\ref{#1}}
\newrefformat{fig}{Fig.\,\ref{#1}}
\newrefformat{tab}{Table\,\ref{#1}}

\newcommand{\bdmath}{\begin{dmath}}
\newcommand{\edmath}{\end{dmath}}
\newcommand{\beq}{\begin{equation}}
\newcommand{\eeq}{\end{equation}}
\newcommand{\bdm}{\begin{displaymath}}
\newcommand{\edm}{\end{displaymath}}
\newcommand{\bea}{\begin{eqnarray}}
\newcommand{\eea}{\end{eqnarray}}
\newcommand{\beal}{\beq \begin{array}{ll}}
\newcommand{\eeal}{\end{array} \eeq}
\newcommand{\beas}{\begin{eqnarray*}}
\newcommand{\eeas}{\end{eqnarray*}}
\newcommand{\ba}{\begin{array}}
\newcommand{\ea}{\end{array}}
\newcommand{\bit}{\begin{itemize}}
\newcommand{\eit}{\end{itemize}}
\newcommand{\ben}{\begin{enumerate}}
\newcommand{\een}{\end{enumerate}}

\newcommand{\hide}[1]{}

\newcommand{\hiddenText}{{\color{gray} hidden text.}}
\newcommand{\hideWithText}[1]{\hiddenText}

\newcommand{\blue}[1]{{\color{blue}#1}}

\newcommand{\linkToPdf}[1]{\href{#1}{\blue{(pdf)}}}
\newcommand{\linkToPpt}[1]{\href{#1}{\blue{(ppt)}}}
\newcommand{\linkToCode}[1]{\href{#1}{\blue{(code)}}}
\newcommand{\linkToWeb}[1]{\href{#1}{\blue{(web)}}}
\newcommand{\linkToVideo}[1]{\href{#1}{\blue{(video)}}}
\newcommand{\linkToMedia}[1]{\href{#1}{\blue{(media)}}}
\newcommand{\award}[1]{\xspace} 

\newacronym{acr:mapf}{MAPF}{Multi-Agent Path Finding}

\newcommand{\myparagraph}[1]{\noindent\textbf{#1}}
\newcommand*\circled[1]{\tikz[baseline=(char.base)]{
            \node[shape=circle,draw,inner sep=0.5pt] (char) {#1};}}

\renewcommand{\baselinestretch}{0.98}

\begin{document}

\title{Multi-Agent Path Finding via  \\ Finite-Horizon Hierarchical Factorization}




%
\author{\authorblockN{Jiarui Li\authorrefmark{1},
Alessandro Zanardi\authorrefmark{2}, 
and Gioele Zardini\authorrefmark{1}}
\authorblockA{\authorrefmark{1}Laboratory for Information and Decision Systems, 
Massachusetts Institute of Technology}
\authorblockA{\authorrefmark{2}Embotech AG}
}

\maketitle

\begin{abstract}
We present a novel algorithm for large-scale \gls{acr:mapf} that enables fast, scalable planning in dynamic environments such as automated warehouses.
Our approach introduces finite-horizon hierarchical factorization, a framework that plans one step at a time in a receding-horizon fashion.
Robots first compute individual plans in parallel, and then dynamically group based on spatio-temporal conflicts and reachability.
The framework accounts for conflict resolution, and for immediate execution and concurrent planning, significantly reducing response time compared to offline algorithms.
Experimental results on benchmark maps demonstrate that our method achieves up to 60\% reduction in time-to-first-action while consistently delivering high-quality solutions, outperforming state-of-the-art offline baselines across a range of problem sizes and planning horizons.
\end{abstract}

\IEEEpeerreviewmaketitle

\section{Introduction}
\label{sec:introduction}


Modern logistics increasingly relies on large-scale automated warehouses and fulfillment centers, where thousands of mobile robots must \emph{coordinate} in shared spaces to fulfill tasks efficiently.
This paradigm shift promises transformative gains in throughput and flexibility, but also presents significant challenges~\cite{wurman2008coordinating, standley2010finding}. 
One of the core difficulties lies in \gls{acr:mapf}, where robots must navigate from start to goal location on a network without colliding.
The complexity of such planning problems grows \emph{exponentially} with the number of agents~\cite{silver2005cooperative}, making even near-optimal solutions intractable at scale.
In standard \gls{acr:mapf}, each robot must avoid \emph{vertex} and \emph{edge} conflicts, i.e., occupying the same vertex or traversing the same edge simultaneously with another robot~\cite{stern2019mapf}.

\myparagraph{Existing Approaches} --
Numerous \gls{acr:mapf} algorithms have been developed, but most are limited in ways that hinder their use in large-scale, real-time warehouse operations.
A critical limitation is that many of these algorithms are \emph{offline}, i.e., they require a complete, globally consistent plan before any robot can begin moving.
This design prevents early execution of even the first steps, as plans may change due to backtracking, introducing significant delays in time-sensitive environments.
\gls{acr:mapf} algorithms also vary in their \emph{performance guarantees}.
Algorithms such as CBS~\cite{sharon2015conflict}, EECBS~\cite{li2021eecbs}, and M*~\cite{wagner2015subdimensional} provide optimal or bounded-suboptimal solutions but scale poorly due to the exponential growth of the joint search space and the inherent NP-hardness~\cite{yu2013planning}.
Faster alternatives, such as MAPF-LNS2~\cite{li2022mapf}, LaCAM~\cite{okumura2023lacam}, and anytime variants such as MAPF-LNS~\cite{li2021anytime}, LaCAM*~\cite{okumura2023improving}, and engineered LaCAM*~\cite{okumura2023engineering}, improve feasibility but still suffer from offline computation and may yield subpar solution quality~\cite{shen2023tracking}.
Furthermore, despite recent progress~\cite{lee2021parallel, okumura2023engineering,chan2024anytime}, most \gls{acr:mapf} planners are sequential, underutilizing modern parallelization techniques.
These limitations motivate the need for \emph{online}, \emph{scalable}, and \emph{parallelizable} planning tools to support real-time robot coordination at scale.

\begin{figure}[t]
    \centering
    \includegraphics[width=\linewidth, trim=1.7cm 5.5cm 1.75cm 5.5cm, clip]{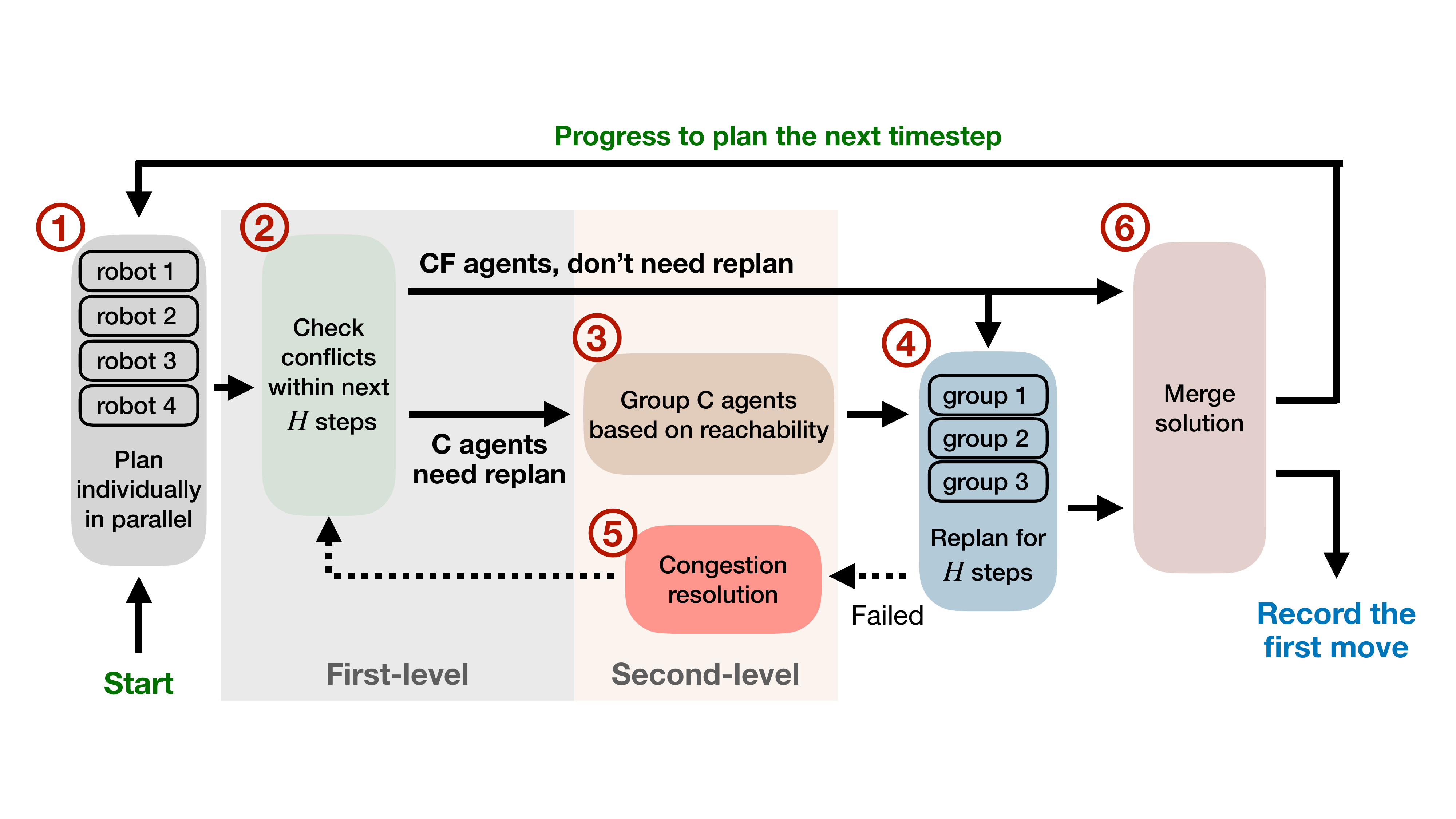}
    \caption{The algorithm plans in a receding-horizon fashion, computing and finalizing robot movements one timestep at a time.
    This online approach enables immediate execution of each computed step.}
    \label{fig:finite_horizon_hierarchical_factorization}
\end{figure}


\myparagraph{Factorization for scalable multi-agent planning} -- 
\emph{Factorization} has emerged as a powerful tool across domains such as game theory~\cite{zanardi2022factorization} and motion planning~\cite{zanardi2023factorization} to mitigate the curse of dimensionality by exploiting the \emph{problem structure}.
The core idea is to leverage \emph{compositionality}\footnote{I.e., the ``solution of the composition of problems'' is the ``composition of the solutions of problems''.}{}: if subgroups of agents can be planned independently, their solutions can be composed, preserving feasibility and often optimality, withing the larger system.
In \gls{acr:mapf}, prior methods have pursued this via bottom-up strategies: assuming full independence, then merging agents that conflict~\cite{wagner2015subdimensional,standley2010finding,lee2021parallel}.
However, because these approaches consider the entire time horizon at once, independence is rare.
As a result, agents are frequently replanned in overlapping groups, limiting scalability.


\myparagraph{Contribution} --
We propose a novel online algorithm for large-scale \gls{acr:mapf} that combines finite-horizon hierarchical factorization with highly parallelizable planning.
The proposed method leverages a uniform greedy optimal planner for individual agents and dynamically groups robots based on conflict and reachability within a finite horizon.
Conflict resolution is handled efficiently using a PIBT-based, state-of-the-art routine~\cite{okumura2022priority}.
Our method enables immediate execution of each computed step, significantly reducing delay compared to offline methods.
Experiments demonstrate faster execution onset and competitive solution quality across diverse scenarios.

\section{Finite-Horizon Hierarchical Factorization}
\label{sec:finite_horizon_hierarchical_factorization}
\cref{fig:finite_horizon_hierarchical_factorization} illustrates the presented algorithm in parallelized stages.
At each iteration, all robots first compute individual paths in parallel leveraging a balanced greedy planner based on backward BFS~\cite{okumura2022priority} (\circled{1}).
Conflicts over the next~$H$ timesteps are then detected via spatial hashing, enabling a first-level factorization: conflict-free robots are finalized, while the rest replan while treating finalized robots as dynamic obstacles (\circled{2}).
Next, conflicting robots are recursively grouped based on horizon-limited reachability (\circled{3}) and replanned in parallel using an adapted PIBT algorithm (\circled{4}).
If replanning fails, a congestion resolution module enlarges the group to allow for more spatio-temporal flexibility (\circled{5}).
Finally, trajectories are merged, and only the first step is executed (\circled{6}).

The proposed algorithm is fast because both individual planning and groupwise replanning are parallelizable.
In practice, many robots remain conflict-free within each horizon, allowing them to follow near-optimal paths while groupwise replanning ensures overall feasibility.
This yields solution quality competitive with leading offline methods and significantly outperforms LaCAM*~\cite{okumura2023improving}, the only known offline planner with comparable speed.

\section{Experimental Results}
\label{sec:experiment_results}
%
\myparagraph{Experiment Setup} -- 
The experiments are conducted using \gls{acr:mapf} benchmarks~\cite{stern2019mapf}\footnote{The experiments were performed on a MacBook Pro 2023 with a 12-core CPU and 36 GB of RAM. The parallel computation used 12 threads.
A video of the experiments is available at \href{https://youtu.be/v3HfqYDTkGY}{youtu.be/v3HfqYDTkGY}.}{}.
Due to space constraints, we report results on two challenging maps: the largest warehouse map \texttt{warehouse-20-40-10-2-2} and the random map \texttt{random-64-64-10}. 
Each experiment was run 200 times, and we report average performance with statistical attributes.
%

\begin{figure}[t]
    \centering
    \includegraphics[width=\linewidth, trim=0.3cm 0.35cm 0cm 0.25cm, clip]{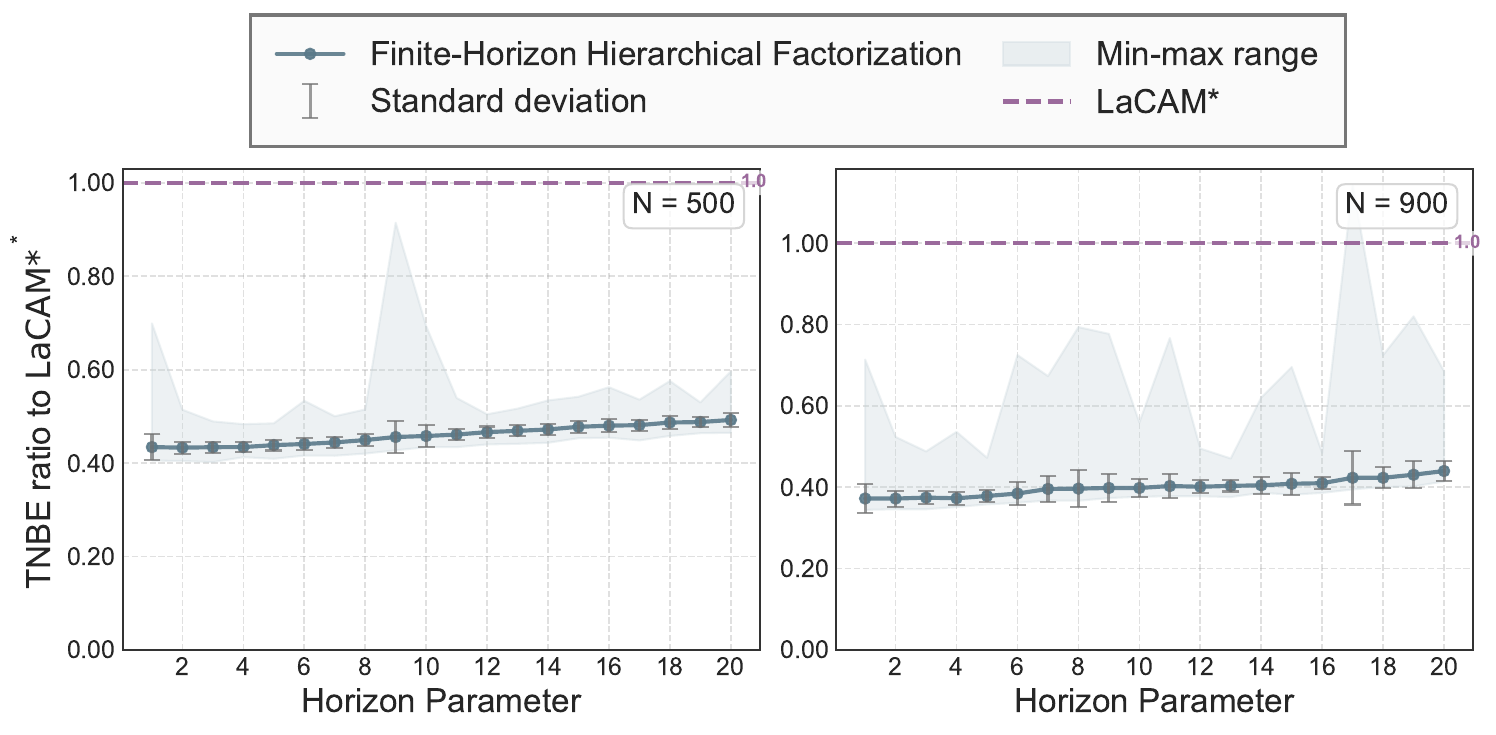}
    \caption{The ratio of the TNBE between our algorithm and offline baselines; values below 1 indicate that our method achieves faster execution readiness.}
    \label{fig:time_before_exe_warehouse}
\end{figure}

%
\myparagraph{Time needed before execution} --
To quantify the advantage of online algorithms, we report the time needed before execution (TNBE).
\cref{fig:time_before_exe_warehouse} shows the TNBE ratio between our method and the LaCAM* baseline.
While TNBE naturally increases with the number of robots, our algorithm consistently achieves substantial speedups, even at the longest planning horizon (20 steps).
For instance, with 900 robots, we observe a 60\% reduction in TNBE.
Moreover, since each planning step in the proposed method completes in under 30 ms, even at scale, subsequent steps can be planned in parallel with execution, significantly outpacing real-world robot actuation times.

\begin{figure}[t]
    \centering
    \includegraphics[width=\linewidth, trim=0.3cm 0.35cm 0cm 0.25cm, clip]{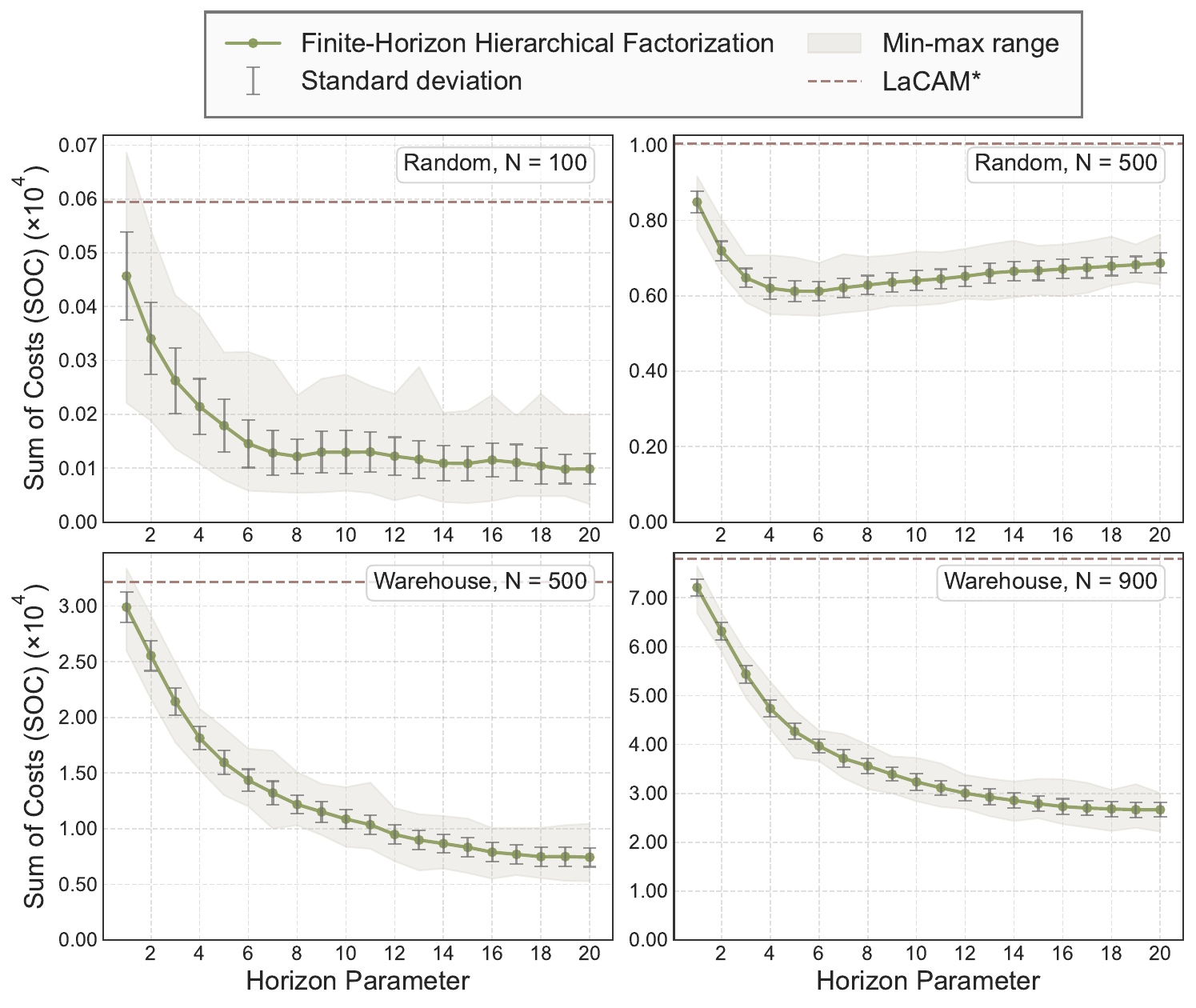}
    \caption{
    Our algorithm consistently outperforms LaCAM* in solution quality across all robot counts. In the warehouse map, performance improves monotonically as the planning horizon increases, while in the random map, the optimal planning horizon varies depending on the number of robots.}
    \label{fig:performance_experiments}
\end{figure}

\myparagraph{Solution quality} --
Solution quality comparisons with LaCAM* are shown in \cref{fig:performance_experiments}, using the sum of costs (SOC), i.e., the total travel time of all agents, measured relative to a lower bound that ignores conflicts~\cite{okumura2023improving}.
In our setup, LaCAM* is allowed to continue refining its solution until our algorithm completes all planning steps.
On the warehouse map, our method consistently yields higher-quality solutions across all robot counts and horizons.
Longer horizons further improve performance by enabling better coordination.
On the random map, our algorithm also outperforms LaCAM*, though gains are not strictly monotonic in horizon length: excessively long horizons reduce the number of conflict-free agents, leading to more frequent and suboptimal groupwise replanning.


\section{Discussion and Conclusion}
\label{sec:discussion_and_conclusion}
We introduced a scalable online algorithm for \gls{acr:mapf} based on finite-horizon hierarchical factorization.
By combining parallel individual planning with dynamic two-level grouping based on conflicts and reachability, as well as efficient on-the-fly conflict resolution, our method enables immediate execution and significantly reduces planning delays.
Experiments confirm its strong performance both in speed and solution quality, making it a practical alternative to existing offline methods for real-time multi-robot coordination.
Future work includes extending the approach to more general task assignment settings~\cite{zardini2022analysis}, lifelong MAPF~\cite{ma2017lifelong}, co-design with networks~\cite{9963724}, learning-enabled improvements~\cite{tresca2025robo}, and tighter guarantees.



\myparagraph{Acknowledgments} --
This work was supported by Prof. Zardini's grant from the MIT Amazon Science Hub, hosted in the Schwarzman College of Computing. 
We thank Dr. Federico Pecora, from Amazon Robotics, Movement Science, for the fruitful discussions and feedback.

\bibliography{references}

\end{document}